\documentclass{article} 
\usepackage{iclr2025_conference,times}

\usepackage{amsmath,amsfonts,bm}

\def\eqref#1{equation~\ref{#1}}

\def\1{\bm{1}}

\DeclareMathAlphabet{\mathsfit}{\encodingdefault}{\sfdefault}{m}{sl}
\SetMathAlphabet{\mathsfit}{bold}{\encodingdefault}{\sfdefault}{bx}{n}

\newcommand{\R}{\mathbb{R}}

\usepackage{hyperref}
\usepackage{url}
\usepackage{graphicx}
\usepackage{tikz}
\usepackage{wrapfig}
\usepackage{mathtools}

\title{Shape Generation Via Weight Space Learning}

\author{Maximilian Plattner, Arturs Berzins, Johannes Brandstetter\thanks{Emmi AI GmbH, Linz, Austria} \\
ELLIS Unit, LIT AI Lab, Institute for Machine Learning, JKU Linz, Austria \\
\texttt{\{plattner, berzins, brandstetter\}@ml.jku.at}
}

\iclrfinalcopy 
\begin{document}

\maketitle

\begin{abstract}
Foundation models for 3D shape generation have recently shown a remarkable capacity to encode rich geometric priors across both global and local dimensions. 
However, leveraging these priors for downstream tasks can be challenging as real‐world data are often scarce or noisy, and traditional fine‐tuning can lead to catastrophic forgetting.
In this work, we treat the \emph{weight space} of a 
large 3D shape‐generative model as a \emph{data modality} that can be explored directly. We hypothesize that submanifolds within this high‐dimensional weight space can modulate topological properties or fine‐grained part features separately, demonstrating early‐stage evidence via two experiments.
First, we observe a sharp \emph{phase transition} in global connectivity when interpolating in conditioning space, suggesting that small changes in weight space can drastically alter topology.
Second, we show that low‐dimensional reparameterizations yield controlled local geometry changes even with very limited data.
These results highlight the potential of weight space learning to unlock new approaches for 3D shape generation and specialized fine‐tuning.
\end{abstract}

\section{Introduction}

\textbf{3D shape-generative foundation models} have recently emerged as powerful architectures that encode substantial \emph{geometric priors} in their learned weights \citet{sanghi2024waveletlatentdiffusionwala, zhang2024lagemlargegeometrymodel}. These priors operate at both \emph{global} and \emph{local} levels: they capture topological properties (e.g.\ connectivity, number of holes) while also expressing fine‐grained part structure. 
Yet, in real‐world engineering or industrial scenarios, the data used to drive downstream tasks are often not only scarce but also noisy or unevenly sampled (e.g.\ raw scans of physical objects) \citet{berzins2024geometryinformedneuralnetworks}. Consequently, naive fine‐tuning of these large generative models to handle specialized 3D tasks risks catastrophic forgetting or mode collapse~\citet{hu2021loralowrankadaptationlarge, bommasani2022opportunitiesrisksfoundationmodels}.

\begin{figure}[h!]
    \centering
    \includegraphics[width=1.0\textwidth]{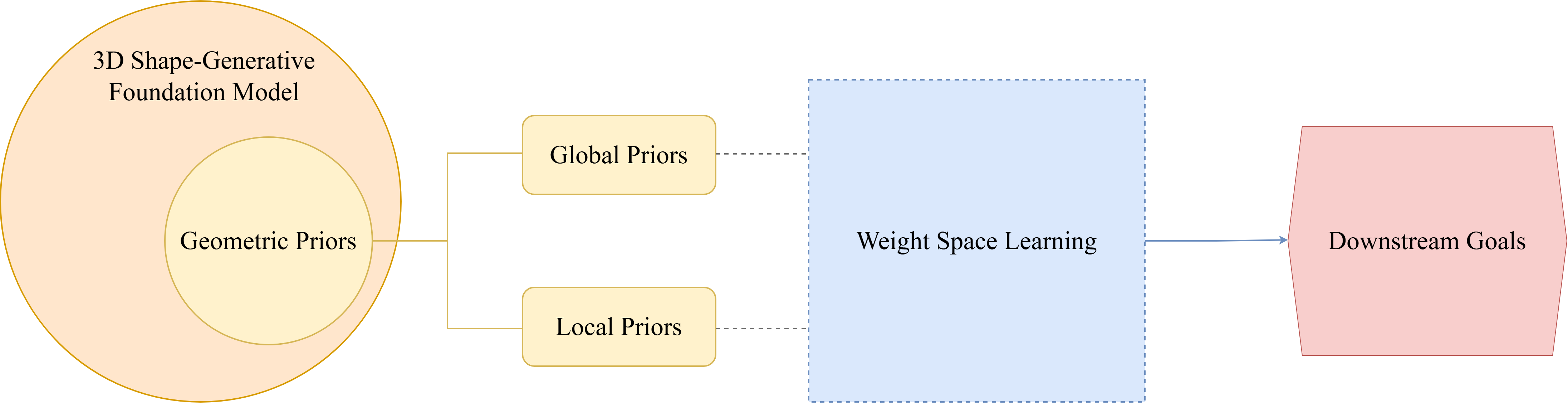}
    \vspace{-1em}
    \caption{\textbf{Conceptual overview.} 3D shape-generative foundation models encode geometric priors (both global and local) in their weights. We hypothesize that these can be leveraged through \emph{weight space learning} in a controlled fashion to address downstream goals.}
    \label{fig:concept_figure}
    \vspace{-0.5em}
\end{figure}

\paragraph{Research question.}
As illustrated in Figure \ref{fig:concept_figure}, we ask how best to reuse the \emph{weight space} of 3D shape-generative foundation models to generate new shapes. 
In particular, we examine through a series of experiments whether small, targeted adjustments in the model’s high‐dimensional weight space can provide handles on both global topological and local geometric features.
\medskip
\paragraph{Results.} We present two complementary results:
\begin{itemize}
    \item \textbf{Global priors:} Subtle movements in weight space can induce sharp \emph{phase transitions} in topological properties. Specifically, we show that even small interpolations in the weight space can abruptly change a shape from one to many connected components.
    \item \textbf{Local priors:} Minor adjustments in weight space can induce focused \emph{local alterations} while maintaining global structural integrity. In particular, we demonstrate that sampling weights from a \emph{submanifold} yields a variety of shapes that preserve the core topology yet express unique local refinements.
\end{itemize}
These results suggest that a deeper understanding of the weight space of 3D shape-generative foundation models may lead to more robust solutions and insights for generative 3D tasks.

\section{Shape Generation via Weight Space Learning}
\label{sec:formalization}

We consider a 3D shape-generative foundation model $f$, parameterized by a high-dimensional weight vector \(\boldsymbol{\theta} \in \mathbb{R}^W\). In practice, \(f\) might be a conditional neural field \citet{xie2022neuralfieldsvisualcomputing} or a diffusion-based pipeline \citet{rombach2022highresolutionimagesynthesislatent}, but we remain agnostic to the specific design choices. Given an input \(\boldsymbol{x} \in \mathcal{X}\) (a noise vector or latent seed), the model produces a shape \(\boldsymbol{s} \in \mathcal{S}\):
\[
  f\bigl(\boldsymbol{x}; \boldsymbol{\theta}\bigr) \;\mapsto\; \boldsymbol{s},
\]
where \(\boldsymbol{s}\) may be represented as a mesh, signed distance function, or another geometric representation. In many models, one can optionally provide a condition \(\boldsymbol{c} \in \mathbb{R}^{C}\) (with \(C \ll W\)) to reconfigure the model’s effective weights $\boldsymbol{\theta}_{\boldsymbol{c}} \in \mathbb{R}^W$ via a map
\[
  \Phi : \mathbb{R}^C \;\to\; \mathbb{R}^W,
  \quad
  \boldsymbol{\theta}_{\boldsymbol{c}} \;=\; \Phi\bigl(\boldsymbol{c}\bigr).
\]
Since \(\Phi\) is continuous, a \(d\)-dimensional manifold in \(\mathbb{R}^C\) induces a corresponding manifold in \(\mathbb{R}^W\)---small shifts in \(\boldsymbol{c}\) lead to coherent reconfiguration of \(\boldsymbol{\theta}\). In a one-dimensional interpolation (i.e.\ \(d=1\)), this amounts to tracing a continuous curve through \(\mathbb{R}^W\)~\citet{rebain2022attentionbeatsconcatenationconditioning}. Crucially, the training process endows \(\boldsymbol{\theta}\) with powerful geometric priors \citet{dupont2022datafunctadatapoint}, capturing both global topological attributes and local shape features.

\begin{figure}[h!]
    \centering
    \includegraphics[width=0.8\textwidth]{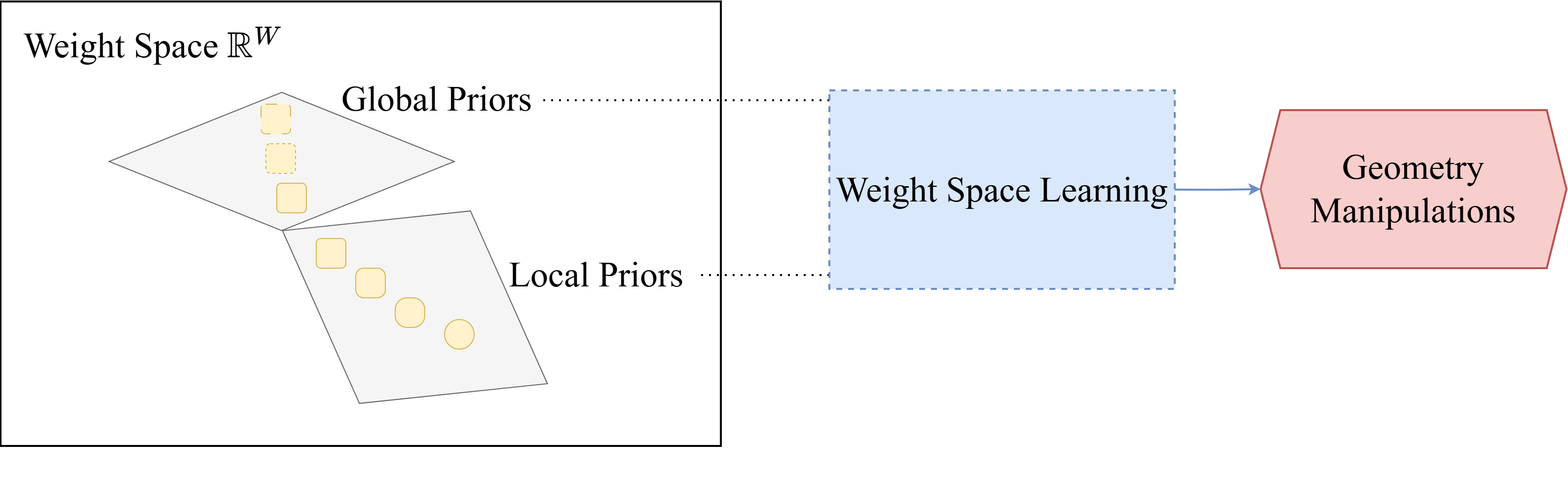}
    \vspace{-0.8em}
    \caption{\textbf{Hypothesis (schematic).} Structured modulations of the weight space $\mathbb{R}^W$ can \textit{isolate} global vs.\ local shape changes and can be \emph{learned}, enabling targeted geometry manipulations.}
    \label{fig:schematic_figure}
    \vspace{-1em}
\end{figure}

\paragraph{Hypothesis (informal).}
As depicted in Figure \ref{fig:schematic_figure}, we posit that within the weight space $\R^W$, submanifolds \textit{exist} that selectively modulate global vs.\ local shape attributes. Moreover, these submanifolds are \textit{learnable}. More precisely:
\begin{enumerate}
    \item \textbf{Existence}: There are weight-space regions that isolate global topology (e.g., connectivity) or local geometry (e.g., part details).
    \item \textbf{Learnability}: These manifolds can be discovered or approximated with common learning techniques.
\end{enumerate}
Hence, by navigating $\mathbb{R}^W$ \emph{directly}, one 
can induce controlled geometry‐aware manipulations. This would prove particularly valuable for downstream tasks with limited or noisy examples. In Section \ref{sec:results}, we present two early-stage experiments that substantiate this premise.

\medskip
\section{Results}
\label{sec:results}
 For our experiments, we consider a 3D shape-generative foundation model
$f\bigl(\boldsymbol{x}; \boldsymbol{\theta}\bigr) \;\mapsto\; \boldsymbol{s}$,
where \(\boldsymbol{\theta} \in \mathbb{R}^W\) with \(W \approx 10^9\) parameters.
Specifically, we use the latent-diffusion model proposed by \citet{sanghi2024waveletlatentdiffusionwala},
trained on 10 million shapes.
The model takes a random noise input \(\boldsymbol{x}\) and a point cloud that is encoded to produce a condition \(\boldsymbol{c} \in \mathbb{R}^C\) with $C=1024^2$. The output $\boldsymbol{s}$ is a signed distance function. Further experiment details can be found in Appendix \ref{appendix:experiment_details}.

\subsection{Global Priors}
\label{sec:global_priors}

\begin{figure}[h!]
    \centering
    \includegraphics[width=0.8\textwidth]{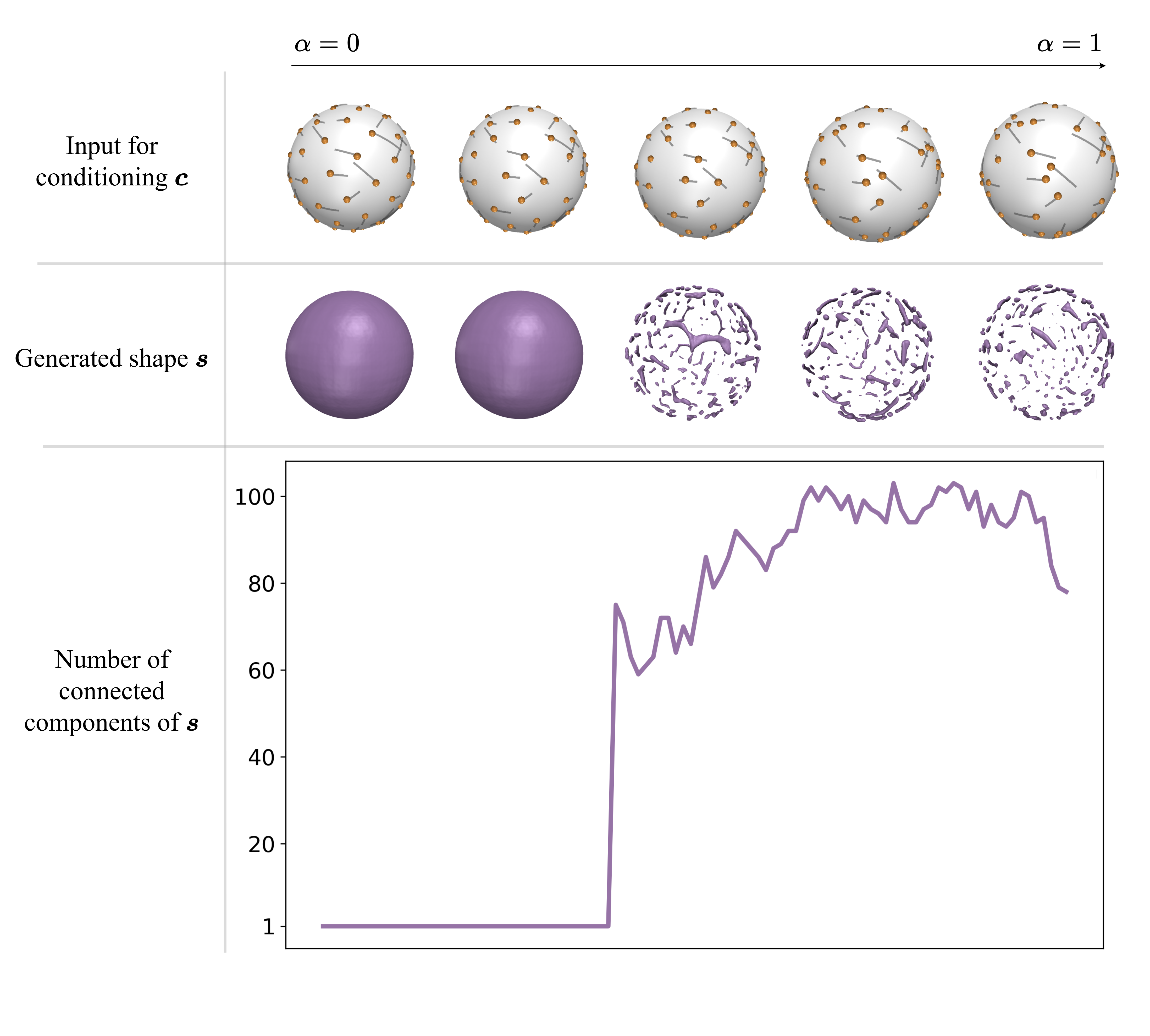}
    \caption{\textbf{Phase transition in connectivity.} 
    By interpolating between evenly and unevenly sampled point clouds on the unit sphere (top row) from $\alpha=0$ to $\alpha=1$, we observe a sudden jump in the generated shapes (middle row) from a single connected component to many disconnected parts, indicating that minimal changes in weight space can induce abrupt global topological shifts (bottom plot).
}
    \label{fig:global_priors_figure}
\end{figure}

In many industrial or engineering contexts, scanning is used to capture real 3D shapes, producing point clouds that are often noisy or uneven -- in contrast to the synthetic data on which 3D shape models are typically trained. Consequently, any fragmentation in the final shape reflects how the reconstruction algorithm, here \(f\),  infers geometry from these scans. We therefore investigate whether minimal shifts in the model's weight space can markedly alter shape connectivity. As shown in Figure~\ref{fig:global_priors_figure}, even small perturbations in $\mathbb{R}^W$ can trigger sharp transitions from a single connected component to multiple fragments.

We create two endpoint point clouds on a unit sphere: one evenly sampled and the other formed by adding noise along the spherical surface to induce unevenness. We then perform a spherical interpolation of these endpoints as $\alpha$ ranges from 0 to 1 in small increments.
At each $\alpha$, we obtain the condition $\boldsymbol{c}$, decode the resulting shape $\boldsymbol{s}$ and measure its number of connected components.

We observe a sharp jump from one to many components at a critical $\alpha^\star$.
This abrupt transition indicates that the foundation model encodes sudden topological phase changes within a narrow region of weight space.

\subsection{Local Priors}
\label{sec:local_priors}

\begin{figure}[h!]
    \centering
\includegraphics[width=0.8\textwidth]{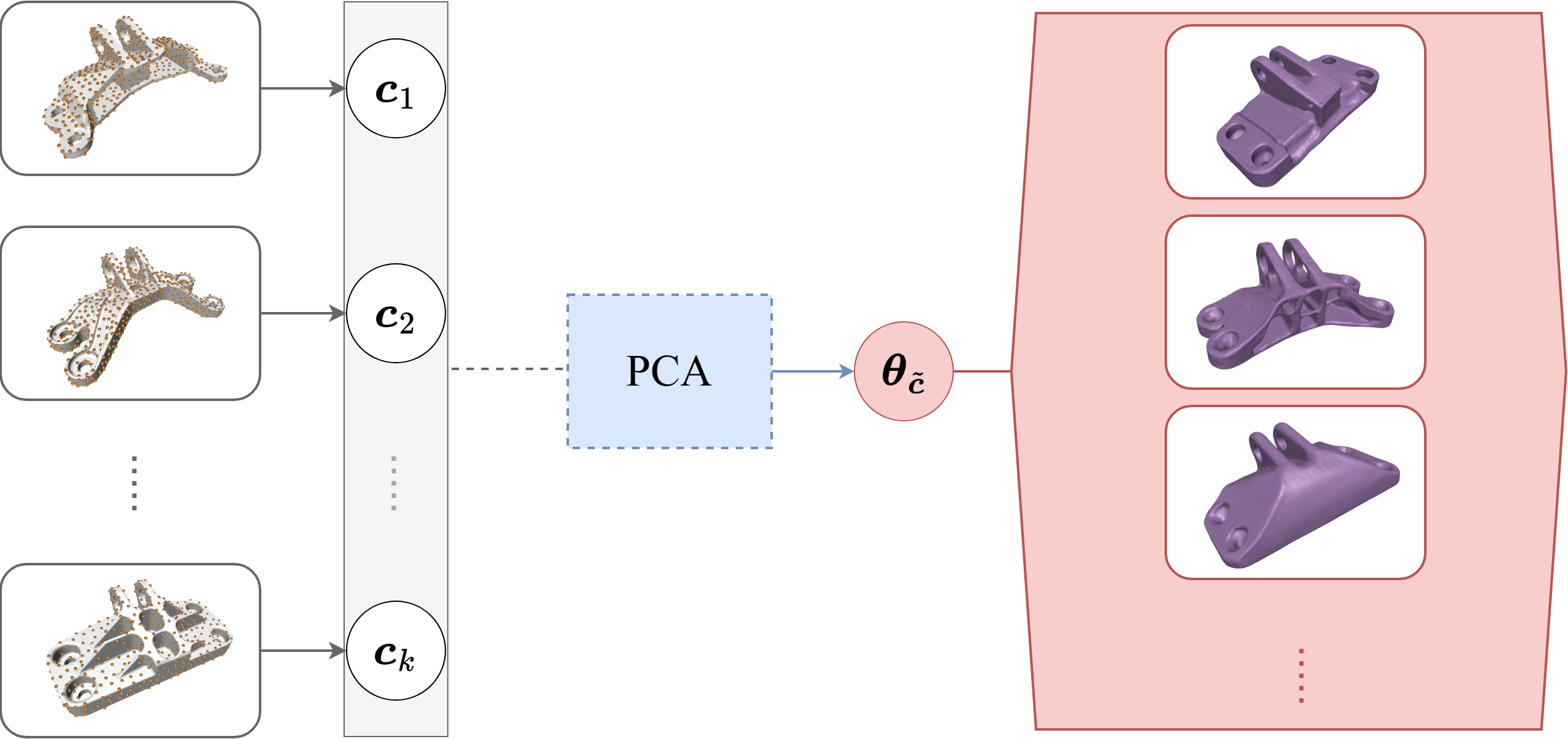}
    \caption{\textbf{Generative modeling of scarce data (SimJEB).} 
   We sample point clouds from $k=381$ unseen SimJEB bracket shapes to obtain conditions $\mathbf{c}_i$ (left). We extract principal components, and sample new weights $\boldsymbol{\theta}_{\tilde{\mathbf{c}}}$ within this PCA subspace (center). Ultimately, we can produce novel bracket designs that preserve overall topology yet exhibit distinct local geometry (right). 
}
    \label{fig:local_priors_figure}
\end{figure}

We address shape generation in data-scarce regimes, where full fine-tuning often leads to overfitting \citet{hu2021loralowrankadaptationlarge}. To that end, we use the SimJEB dataset \citet{Whalen_2021} of $k=381$ mechanical bracket shapes, which our model $f$ has not seen during pre-training. As outlined in Figure \ref{fig:local_priors_figure}, we assess the ability of the weight space to preserve global topology while enabling fine-grained geometry changes.

For each of the bracket shapes in SimJEB, we sample surface points to obtain conditionings  \(\{\mathbf{c}_i\}_{i=1}^k\). We then compute their sample covariance, and extract principal components. We generate new weights $\boldsymbol{\theta}_{\tilde{\boldsymbol{c}}}$ by sampling within this PCA-derived subspace and inspect the generated shapes $\boldsymbol{s}$.

Empirically, random or interpolated weights within this submanifold yield bracket designs that remain structurally coherent yet exhibit distinct local details.
This finding demonstrates the capacity of the weight space for local geometric refinements. In data-scarce environments, this further allows us to generate diverse shapes without fine-tuning $f$.

\medskip

\section{Discussion}
\vspace{-0.5em}
Our experiments show that small changes in the conditioning space and thus in the underlying weight configuration of the model cause controlled changes in the generated 3D shapes. In the global priors setting, this was evident in the abrupt phase transition of connectivity, while for local priors, low‐dimensional manipulations in weight space introduced fine‐grained geometric variations.

\paragraph{Limitations and future work.}
Building on these findings, a natural next step is to investigate the mechanisms behind the high topological sensitivity. A broader comparative study of different architectures
(e.g.\ neural fields vs.\ diffusion) and training regimes could clarify whether these behaviors generalize or
arise from particular modeling choices.
By refining our understanding of how tiny perturbations in parameter space
translate to controlled geometric changes, we can move toward building more predictable, robust pipelines for
3D shape generation. We hope this work encourages further research at the intersection of foundation models, geometry, and weight space analysis.

\newpage
\bibliography{iclr2025_conference}

\begin{thebibliography}{10}
\providecommand{\natexlab}[1]{#1}
\providecommand{\url}[1]{\texttt{#1}}
\expandafter\ifx\csname urlstyle\endcsname\relax
  \providecommand{\doi}[1]{doi: #1}\else
  \providecommand{\doi}{doi: \begingroup \urlstyle{rm}\Url}\fi

\bibitem[Berzins et~al.(2024)Berzins, Radler, Volkmann, Sanokowski, Hochreiter, and Brandstetter]{berzins2024geometryinformedneuralnetworks}
Arturs Berzins, Andreas Radler, Eric Volkmann, Sebastian Sanokowski, Sepp Hochreiter, and Johannes Brandstetter.
\newblock Geometry-informed neural networks, 2024.
\newblock URL \url{https://arxiv.org/abs/2402.14009}.

\bibitem[Bommasani et~al.(2022)Bommasani, Hudson, Adeli, Altman, Arora, von Arx, Bernstein, Bohg, Bosselut, Brunskill, Brynjolfsson, Buch, Card, Castellon, Chatterji, Chen, Creel, Davis, Demszky, Donahue, Doumbouya, Durmus, Ermon, Etchemendy, Ethayarajh, Fei-Fei, Finn, Gale, Gillespie, Goel, Goodman, Grossman, Guha, Hashimoto, Henderson, Hewitt, Ho, Hong, Hsu, Huang, Icard, Jain, Jurafsky, Kalluri, Karamcheti, Keeling, Khani, Khattab, Koh, Krass, Krishna, Kuditipudi, Kumar, Ladhak, Lee, Lee, Leskovec, Levent, Li, Li, Ma, Malik, Manning, Mirchandani, Mitchell, Munyikwa, Nair, Narayan, Narayanan, Newman, Nie, Niebles, Nilforoshan, Nyarko, Ogut, Orr, Papadimitriou, Park, Piech, Portelance, Potts, Raghunathan, Reich, Ren, Rong, Roohani, Ruiz, Ryan, Ré, Sadigh, Sagawa, Santhanam, Shih, Srinivasan, Tamkin, Taori, Thomas, Tramèr, Wang, Wang, Wu, Wu, Wu, Xie, Yasunaga, You, Zaharia, Zhang, Zhang, Zhang, Zhang, Zheng, Zhou, and Liang]{bommasani2022opportunitiesrisksfoundationmodels}
Rishi Bommasani, Drew~A. Hudson, Ehsan Adeli, Russ Altman, Simran Arora, Sydney von Arx, Michael~S. Bernstein, Jeannette Bohg, Antoine Bosselut, Emma Brunskill, Erik Brynjolfsson, Shyamal Buch, Dallas Card, Rodrigo Castellon, Niladri Chatterji, Annie Chen, Kathleen Creel, Jared~Quincy Davis, Dora Demszky, Chris Donahue, Moussa Doumbouya, Esin Durmus, Stefano Ermon, John Etchemendy, Kawin Ethayarajh, Li~Fei-Fei, Chelsea Finn, Trevor Gale, Lauren Gillespie, Karan Goel, Noah Goodman, Shelby Grossman, Neel Guha, Tatsunori Hashimoto, Peter Henderson, John Hewitt, Daniel~E. Ho, Jenny Hong, Kyle Hsu, Jing Huang, Thomas Icard, Saahil Jain, Dan Jurafsky, Pratyusha Kalluri, Siddharth Karamcheti, Geoff Keeling, Fereshte Khani, Omar Khattab, Pang~Wei Koh, Mark Krass, Ranjay Krishna, Rohith Kuditipudi, Ananya Kumar, Faisal Ladhak, Mina Lee, Tony Lee, Jure Leskovec, Isabelle Levent, Xiang~Lisa Li, Xuechen Li, Tengyu Ma, Ali Malik, Christopher~D. Manning, Suvir Mirchandani, Eric Mitchell, Zanele Munyikwa, Suraj Nair,
  Avanika Narayan, Deepak Narayanan, Ben Newman, Allen Nie, Juan~Carlos Niebles, Hamed Nilforoshan, Julian Nyarko, Giray Ogut, Laurel Orr, Isabel Papadimitriou, Joon~Sung Park, Chris Piech, Eva Portelance, Christopher Potts, Aditi Raghunathan, Rob Reich, Hongyu Ren, Frieda Rong, Yusuf Roohani, Camilo Ruiz, Jack Ryan, Christopher Ré, Dorsa Sadigh, Shiori Sagawa, Keshav Santhanam, Andy Shih, Krishnan Srinivasan, Alex Tamkin, Rohan Taori, Armin~W. Thomas, Florian Tramèr, Rose~E. Wang, William Wang, Bohan Wu, Jiajun Wu, Yuhuai Wu, Sang~Michael Xie, Michihiro Yasunaga, Jiaxuan You, Matei Zaharia, Michael Zhang, Tianyi Zhang, Xikun Zhang, Yuhui Zhang, Lucia Zheng, Kaitlyn Zhou, and Percy Liang.
\newblock On the opportunities and risks of foundation models, 2022.
\newblock URL \url{https://arxiv.org/abs/2108.07258}.

\bibitem[Dupont et~al.(2022)Dupont, Kim, Eslami, Rezende, and Rosenbaum]{dupont2022datafunctadatapoint}
Emilien Dupont, Hyunjik Kim, S.~M.~Ali Eslami, Danilo Rezende, and Dan Rosenbaum.
\newblock From data to functa: Your data point is a function and you can treat it like one, 2022.
\newblock URL \url{https://arxiv.org/abs/2201.12204}.

\bibitem[Hu et~al.(2021)Hu, Shen, Wallis, Allen-Zhu, Li, Wang, Wang, and Chen]{hu2021loralowrankadaptationlarge}
Edward~J. Hu, Yelong Shen, Phillip Wallis, Zeyuan Allen-Zhu, Yuanzhi Li, Shean Wang, Lu~Wang, and Weizhu Chen.
\newblock Lora: Low-rank adaptation of large language models, 2021.
\newblock URL \url{https://arxiv.org/abs/2106.09685}.

\bibitem[Rebain et~al.(2022)Rebain, Matthews, Yi, Sharma, Lagun, and Tagliasacchi]{rebain2022attentionbeatsconcatenationconditioning}
Daniel Rebain, Mark~J. Matthews, Kwang~Moo Yi, Gopal Sharma, Dmitry Lagun, and Andrea Tagliasacchi.
\newblock Attention beats concatenation for conditioning neural fields, 2022.
\newblock URL \url{https://arxiv.org/abs/2209.10684}.

\bibitem[Rombach et~al.(2022)Rombach, Blattmann, Lorenz, Esser, and Ommer]{rombach2022highresolutionimagesynthesislatent}
Robin Rombach, Andreas Blattmann, Dominik Lorenz, Patrick Esser, and Björn Ommer.
\newblock High-resolution image synthesis with latent diffusion models, 2022.
\newblock URL \url{https://arxiv.org/abs/2112.10752}.

\bibitem[Sanghi et~al.(2024)Sanghi, Khani, Reddy, Rampini, Cheung, Malekshan, Madan, and Shayani]{sanghi2024waveletlatentdiffusionwala}
Aditya Sanghi, Aliasghar Khani, Pradyumna Reddy, Arianna Rampini, Derek Cheung, Kamal~Rahimi Malekshan, Kanika Madan, and Hooman Shayani.
\newblock Wavelet latent diffusion (wala): Billion-parameter 3d generative model with compact wavelet encodings, 2024.
\newblock URL \url{https://arxiv.org/abs/2411.08017}.

\bibitem[Whalen et~al.(2021)Whalen, Beyene, and Mueller]{Whalen_2021}
E.~Whalen, A.~Beyene, and C.~Mueller.
\newblock Simjeb: Simulated jet engine bracket dataset.
\newblock \emph{Computer Graphics Forum}, 40\penalty0 (5):\penalty0 9–17, August 2021.
\newblock ISSN 1467-8659.
\newblock \doi{10.1111/cgf.14353}.
\newblock URL \url{http://dx.doi.org/10.1111/cgf.14353}.

\bibitem[Xie et~al.(2022)Xie, Takikawa, Saito, Litany, Yan, Khan, Tombari, Tompkin, Sitzmann, and Sridhar]{xie2022neuralfieldsvisualcomputing}
Yiheng Xie, Towaki Takikawa, Shunsuke Saito, Or~Litany, Shiqin Yan, Numair Khan, Federico Tombari, James Tompkin, Vincent Sitzmann, and Srinath Sridhar.
\newblock Neural fields in visual computing and beyond, 2022.
\newblock URL \url{https://arxiv.org/abs/2111.11426}.

\bibitem[Zhang \& Wonka(2024)Zhang and Wonka]{zhang2024lagemlargegeometrymodel}
Biao Zhang and Peter Wonka.
\newblock Lagem: A large geometry model for 3d representation learning and diffusion, 2024.
\newblock URL \url{https://arxiv.org/abs/2410.01295}.

\end{thebibliography}
\bibliographystyle{iclr2025_conference}

\newpage
\appendix

\appendix

\appendix

\section{Experiment Details}
\label{appendix:experiment_details}

In this section, we provide additional information on the experimental setup for both the \emph{Global Priors}
(\S\ref{sec:global_priors}) and the \emph{Local Priors} (\S\ref{sec:local_priors}) experiments.
Our goal is to ensure reproducibility without excessive low‐level specifics.

\subsection{Global Priors (Connectivity)}
\label{appendix:global_priors}

\paragraph{Model and inputs.}
All global priors experiments rely on the same point-cloud-conditional latent diffusion model \citet{sanghi2024waveletlatentdiffusionwala}, with $W \approx 10^9$. We keep the random noise input \(\boldsymbol{x}\)
fixed by setting a diffusion seed (42 unless stated otherwise) for clarity, but the reported behaviors hold across different seeds.

\paragraph{Fibonacci-sampled base sphere.}
We begin with an \emph{evenly} distributed point cloud \(\mathbf{A}\) on the unit sphere using the
\emph{Fibonacci (Golden Spiral)} method, with $N=1000$ points.

\paragraph{Noisy endpoint.}
To introduce variability, we perturb each point \(\mathbf{A}_i\) by adding a random direction, then
\emph{re‐normalize} to ensure each result remains on the unit sphere:
\[
  \mathbf{B}_i \;=\;\frac{\mathbf{A}_i + \boldsymbol{\delta}_i}{\|\mathbf{A}_i + \boldsymbol{\delta}_i\|}
  \quad\text{with}\quad
  \boldsymbol{\delta}_i \sim \mathcal{N}(0, \sigma^2\mathbf{I}).
\]
This yields a \(\mathbf{B}\) that is also strictly on the sphere, but with an uneven distribution.

\paragraph{Spherical linear interpolation (slerp).}
To traverse a geodesic on the sphere from \(\mathbf{A}_i\) to \(\mathbf{B}_i\), we use a standard
slerp formula for each point. For \(\alpha\in [0,1]\), we compute:
\[
  \mathbf{P}_{\alpha,i} \;=\; \mathrm{slerp}(\mathbf{A}_i, \mathbf{B}_i, \alpha),
\]
preserving unit‐length throughout. Repeating this across all points \(i\) yields the interpolated
point cloud \(\mathbf{P}_\alpha\). We choose a grid of \(\alpha\) values with 500 evenly‐spaced steps from 0 to 1.

\paragraph{Mesh reconstruction \& connectivity.}
We pass each point cloud \(\mathbf{P}_\alpha\) through the model’s encoder to obtain the
conditioning, then decode it into a 3D mesh with the pipeline of \citet{sanghi2024waveletlatentdiffusionwala}.  
Using \texttt{trimesh}, we count the number of connected components by splitting the decoded mesh
into watertight submeshes. As reported in \S\ref{sec:global_priors}, we observe a sudden “phase
transition” at a critical \(\alpha^*\), with the number of components jumping from one to many.

\paragraph{Multiple runs.}
To ensure consistency, we repeat the above procedure across multiple random seeds (for both the
noise \(\boldsymbol{\delta}_i\) and the diffusion model’s random sampling), confirming the
abrupt nature of the global topological change.

\subsection{Local Priors (SimJEB Bracket Shapes)}
\label{appendix:local_priors}

\paragraph{Model and dataset.}
We use the same diffusion‐based foundation model of \citet{sanghi2024waveletlatentdiffusionwala}
as in the global priors experiment, but now condition on point clouds derived from SimJEB \citet{Whalen_2021}.
The SimJEB dataset contains \(k=381\) mechanical bracket CAD models.

\paragraph{Point cloud sampling.}
For each bracket model, we randomly sample $2500$ surface points, 
scaled such that each model fits within a standard bounding box. 
These point clouds are then encoded to produce the conditioning vectors \(\{\boldsymbol{c}_i\}_{i=1}^k\).

\paragraph{PCA subspace.}
We compute the sample covariance matrix of \(\{\boldsymbol{c}_i\}\) (after mean centering). 
Since each \(\boldsymbol{c}_i\) is high‐dimensional with $C=1024^2$, we keep only the top 100 principal components, which
account for the majority of variance. We then generate new conditioning vectors \(\tilde{\boldsymbol{c}}\) by
sampling and interpolating within this 100D PCA subspace.

\paragraph{Shape decoding.}
We feed these new conditioning vectors \(\tilde{\boldsymbol{c}}\) into the model (again with a
fixed diffusion seed of 42 unless otherwise noted) to produce corresponding bracket meshes. As reported in
\S\ref{sec:local_priors}, the generated shapes preserve key global attributes (e.g.\ connectedness, overall
structure) yet exhibit diverse local variations.

\paragraph{Observational criteria.}
To verify that our method yields realistic brackets with consistent global topology,
we visually inspect the decoded meshes for continuous geometry and plausible bracket shape. We also confirm that
minor perturbations in the PCA subspace lead to fine‐grained but coherent changes (e.g.\ hole size, fillet radius, etc.).

\end{document}